\documentclass{article}
\usepackage{iclr2027_conference,times}

\usepackage{amsmath,amssymb,booktabs,graphicx,multirow,xcolor,caption,wrapfig,needspace,float}
\usepackage{hyperref}
\usepackage{url}
\usepackage{placeins}

\hypersetup{
  colorlinks=true,
  citecolor=blue,
  linkcolor=black,
  urlcolor=blue
}

\title{What Should We Ask Next? Retrieval-Aware Question Learning for Interactive ReID}

\iclrfinalcopy
\author{Lyucheng Qian$^{1}$ \quad John Yuehan Zhang$^{2}$ \quad Pingyu Wang$^{1}$}

\fancypagestyle{arxivtitle}{%
  \fancyhf{}
  \fancyfoot[L]{\parbox[b]{0.72\textwidth}{\scriptsize
    \textsuperscript{1}Sichuan University; \textsuperscript{2}University of California, Berkeley.\\
    Corresponding author: Pingyu Wang.}}
  \fancyfoot[C]{\thepage}
}

\begin{document}
\maketitle
\fancyhead{}
\thispagestyle{arxivtitle}

\begin{abstract}
Interactive retrieval with partial evidence constitutes a sequential information-acquisition problem: an agent must choose questions that acquire useful evidence for the next retrieval update. Existing systems train this decision by imitating an offline ordering of candidate QA pairs. However, a question's value depends on the response it elicits and its downstream effect on retrieval. We establish that candidate discriminativeness and perceived usefulness provide weak supervision for this objective, then introduce RAVEL, a retrieval-aware online reinforcement learning framework for interactive person re-identification. RAVEL initializes from supervised question generation, observes the current Top-4 candidates directly, and optimizes the question policy with rank feedback from the full question--answer--retrieval loop. Experiments on Interactive-PEDES show that RAVEL delivers progressively stronger retrieval performance across five interaction rounds. Further analysis shows that RAVEL allocates more of its interaction budget to localized open-ended prompts targeting specific attributes.
\end{abstract}

\section{Introduction}

Interactive person re-identification (ReID) extends conventional person re-identification with iterative question--answer interactions between a questioner and an answerer. We refer to this interactive setting as Interactive ReID (IReID). At each round, the system asks a targeted question, incorporates the processed response into the retrieval text, and updates the gallery ranking. Unlike conventional ReID, which ranks a gallery from a fixed image or text query, IReID is sequential: as the ranking and evidence evolve, the system must decide what to ask next to distinguish the target from visually similar distractors. A question's value is therefore not determined by its wording alone; it depends on the answer it elicits and how that answer changes subsequent retrieval.

Prior work has studied iterative visual search through relative-attribute feedback \citep{kovashka2015whittlesearch}, dialogue-based image retrieval \citep{levy2023chatting}, and question-driven video retrieval \citep{madasu2022vired,liang2023simple}. In IReID, LLaVA-ReID decomposes fine-grained descriptions into candidate question--answer pairs and uses a greedy look-forward strategy to order them by immediate ranking improvement \citep{lu2025llavareid}. Because this order is fixed before deployment, it cannot adapt as answers update the retrieval text and candidate set. This motivates learning questions from feedback in the realized question--answer--retrieval loop.

Our diagnostics examine whether static question judgments and candidate conditioning reveal downstream retrieval utility. Candidate discriminativeness and perceived usefulness are only weakly associated with realized retrieval gain, while visual-token reliance alone does not ensure sensitivity to candidate composition or order. These findings motivate online policy learning from feedback in the actual question--answer--retrieval loop.

Therefore, we propose \textbf{Retrieval-Aware Verbal Evidence Learning (RAVEL)}, an online question-learning framework for interactive person ReID. Its central idea is to treat question generation as sequential evidence acquisition inside the retrieval loop. Rather than assigning a question value before interaction, RAVEL evaluates a question through the answer it elicits and the ranking change produced by the resulting evidence. The question policy is therefore learned from the complete question--answer--retrieval loop, allowing the current candidate set and accumulated dialogue to influence what is asked next.

Concretely, RAVEL retains a frozen IRRA retriever \citep{jiang2023cross} to maintain the candidate ranking, uses supervised fine-tuning to initialize the questioner, and then updates its QLoRA adapter \citep{dettmers2023qlora} with online reinforcement learning from realized rank feedback. At each round, the policy observes the current Top-4 retrieval results directly; the generated question is answered, the cleaned response is appended to the retrieval text, and the retriever updates the ranking. A validity gate removes protocol-violating actions while preserving free-form question generation, so the learned policy can explore answerable evidence-acquisition strategies without relying on a fixed question vocabulary.

\begin{figure}[t]
\centering
\includegraphics[width=0.94\textwidth]{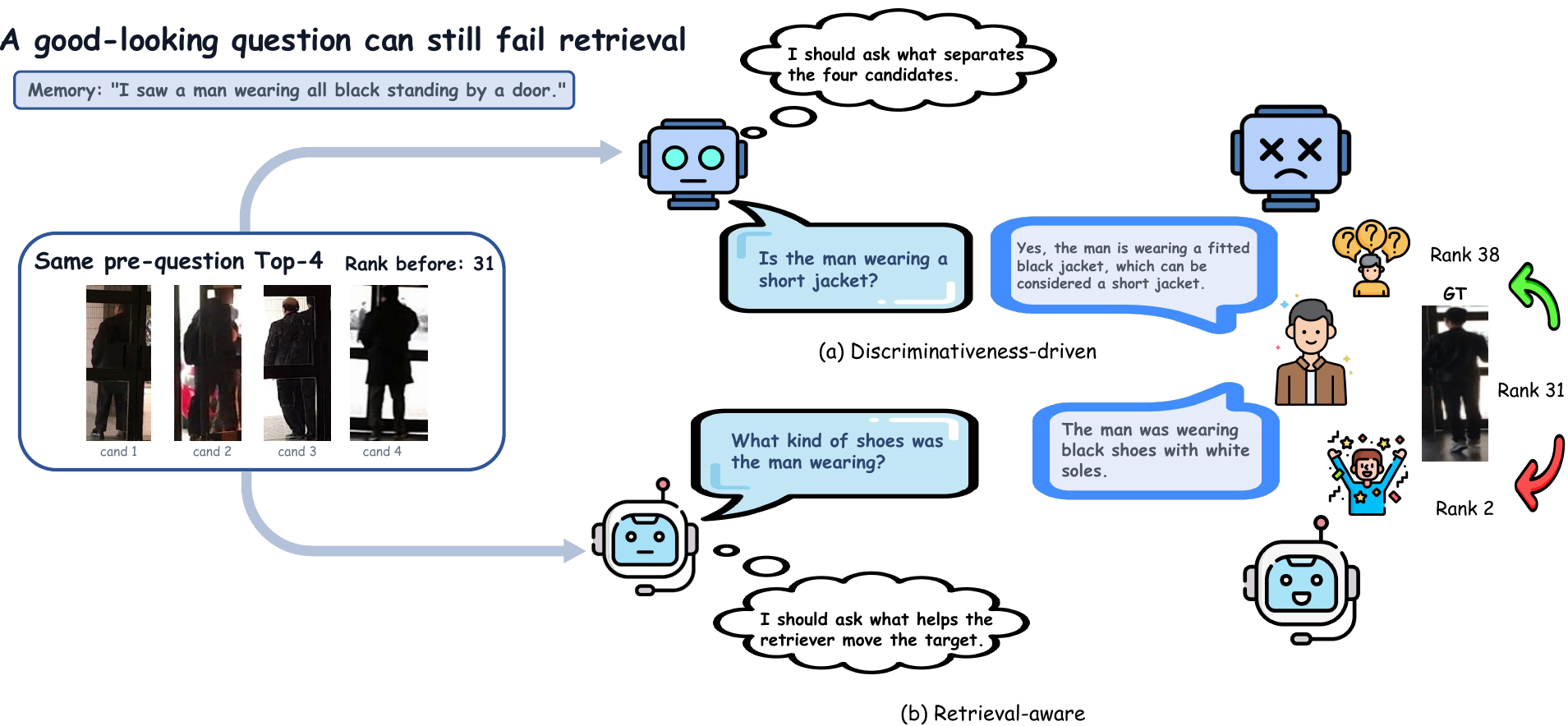}
\caption{Illustrative interactive ReID example. From the same Top-4 state at Rank 31, the jacket question moves the target to Rank 38, whereas the retrieval-aware shoe question raises it to Rank 2; question value follows the evidence elicited and its effect on re-ranking.}
\label{fig:retrieval-aware-case}
\end{figure}

Our contributions are:

1. \textbf{Question-value diagnosis.} We test whether candidate discriminativeness and perceived usefulness predict realized retrieval gains, finding only weak associations. This shows that static judgments alone do not characterize a question's utility after its answer is processed and the gallery is re-ranked.

2. \textbf{Candidate-conditioned visual analysis.} We use attention and likelihood probes to measure how the questioner uses candidate images and responds to candidate composition and order. The results distinguish visual-token reliance from effective candidate sensitivity: attending to images does not by itself ensure questions adapt to the current gallery.

3. \textbf{Retrieval-aware learning and analysis.} We propose online question learning from rank feedback over the complete question--answer--retrieval loop and analyze how the trained policy schedules question types, targets localized evidence, and accumulates retrieval gains.

\section{Related Work}

\subsection{Traditional Person Re-Identification}

Text-based person ReID retrieves a person image from a natural-language description, requiring fine-grained alignment between linguistic attributes and visual evidence \citep{li2017cuhkpedes,zhu2021dssl,ding2021ssan}. Recent cross-modal methods strengthen this matching through IRRA \citep{jiang2023cross}, CFine \citep{yan2023cfine}, RaSa \citep{bai2023rasa}, and AUL \citep{li2024aul}; APTM contributes a large-scale multi-attribute and language retrieval benchmark \citep{yang2023aptm}. Complementary work addresses noisy image--text correspondences and uncertain pedestrian attributes \citep{qin2024rde,sun2026weakpair,lou2026uapar}. However, fixed-query methods cannot select new evidence as candidates evolve; RAVEL adds an online question policy trained from realized rank feedback while retaining the cross-modal retriever.

\subsection{Interactive Person Re-Identification}

Interactive person ReID formulates retrieval as iterative evidence gathering. General interactive retrieval systems such as ChatIR and PlugIR study conversational image retrieval and interactive query refinement before the emergence of Interactive-PEDES \citep{levy2023chatting,lee2024plugir}. LLaVA-ReID later introduced Interactive-PEDES and brought candidate-conditioned question sequencing to interactive person ReID, learning offline question orders from retrieval-based supervision \citep{lu2025llavareid}. Human-centered studies have also explored how multimodal large language models (MLLMs) can assist retrieval interactions; these studies use the model as an auxiliary component and do not learn an interactive ReID question policy \citep{qin2025human}. In adjacent dialog-based image retrieval, \citet{guo2018dialog} optimize per-turn ranking with reinforcement learning. Existing methods therefore cover general interactive retrieval, offline question ordering, or auxiliary MLLM assistance. Free-form question policies trained from answer-conditioned online feedback remain unexplored. RAVEL addresses this gap by updating its question policy online from the complete question--answer--retrieval loop under a fixed interaction budget.

\section{Method}

\begin{figure}[H]
\centering
\includegraphics[width=0.86\textwidth]{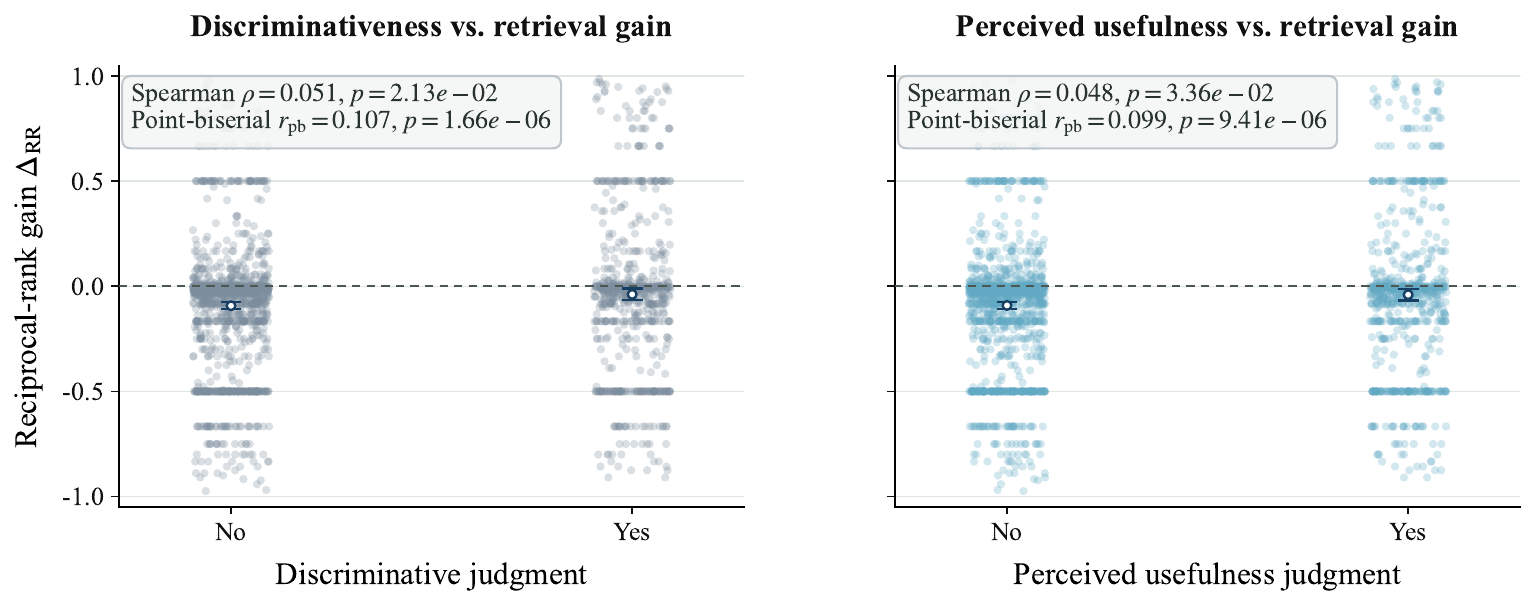}
\caption{Question-quality judgments and downstream retrieval gain, measured by the reciprocal-rank difference $\Delta_{\mathrm{RR}} = 1/r_{\mathrm{after}} - 1/r_{\mathrm{before}}$.}
\label{fig:question-quality}
\end{figure}

\subsection{Motivation}

Let \(M\) denote the private witness memory available only to the answerer, and let $X$ denote the initial description visible to the questioner. At round $t$, $C_t$ is the ordered Top-4 candidate set, $R_t$ is the accumulated retrieval text, and $r_t$ is the target rank maintained by the environment. The environment state is $s_t=(M,X,Q_{<t},A_{<t},C_t,R_t,r_t,t)$, whereas the questioner receives the partial observation $o_t=(X,Q_{<t},A_{<t},C_t,R_t,t)$ and never observes $M$ or $r_t$. The questioner generates $q_t$, the answerer produces $a_t$, and the retriever updates the ranking.
The retrieved text is updated by applying the answer-cleaning rules to the witness response and appending the result to the previous text. We use the same gallery, retriever, answer-length limit, and text preprocessing protocol across comparisons.

\subsubsection{Question-Value Diagnostics}

A natural hypothesis is that a question separating the current Top-4 candidates should be useful. We test it on 2,000 observed test-set interaction states, balanced across rounds, using Qwen3-VL-32B \citep{bai2025qwen3vl} judgments of discriminativeness and perceived usefulness; 101 cases also receive human usefulness labels. The automatic judgments have Spearman correlations \citep{spearman1904} of only $0.051$ and $0.048$ with reciprocal-rank gain, indicating that static assessments barely track the continuous retrieval outcome. Their point-biserial correlations \citep{tate1954} with the rank-improvement event are similarly weak, at $0.107$ and $0.099$, so the conclusion is unchanged when improvement is treated as a binary event. To check whether this pattern is caused by VLM labeling noise, we examine the 101 human labels separately: their correlations with reciprocal-rank gain ($\rho=-0.016$, $p=0.875$) and the improvement event ($r_{\mathrm{pb}}=-0.041$, $p=0.687$) are also not reliable. Human--model agreement is $65.35\%$ with Cohen's $\kappa=0.312$ \citep{cohen1960}, suggesting that the weak association is not explained solely by a mismatch between the automatic and human judgments; the corresponding distributions are shown in Figure~\ref{fig:question-quality}, with label mappings and per-round statistics in Supplementary Material A.

These diagnostics do not evaluate LLaVA-ReID's rank-derived labels; they show that static judgments of question quality and candidate conditioning are weak indicators of retrieval utility after answering and re-ranking. The human subset suggests that this weak association is not solely an artifact of VLM labeling noise. The resulting motivation is to optimize question policies from feedback in the realized question--answer--retrieval loop, where answers update the evidence and candidate state.

\setcounter{figure}{2}
\begin{figure*}[t]
\centering
\includegraphics[width=\textwidth]{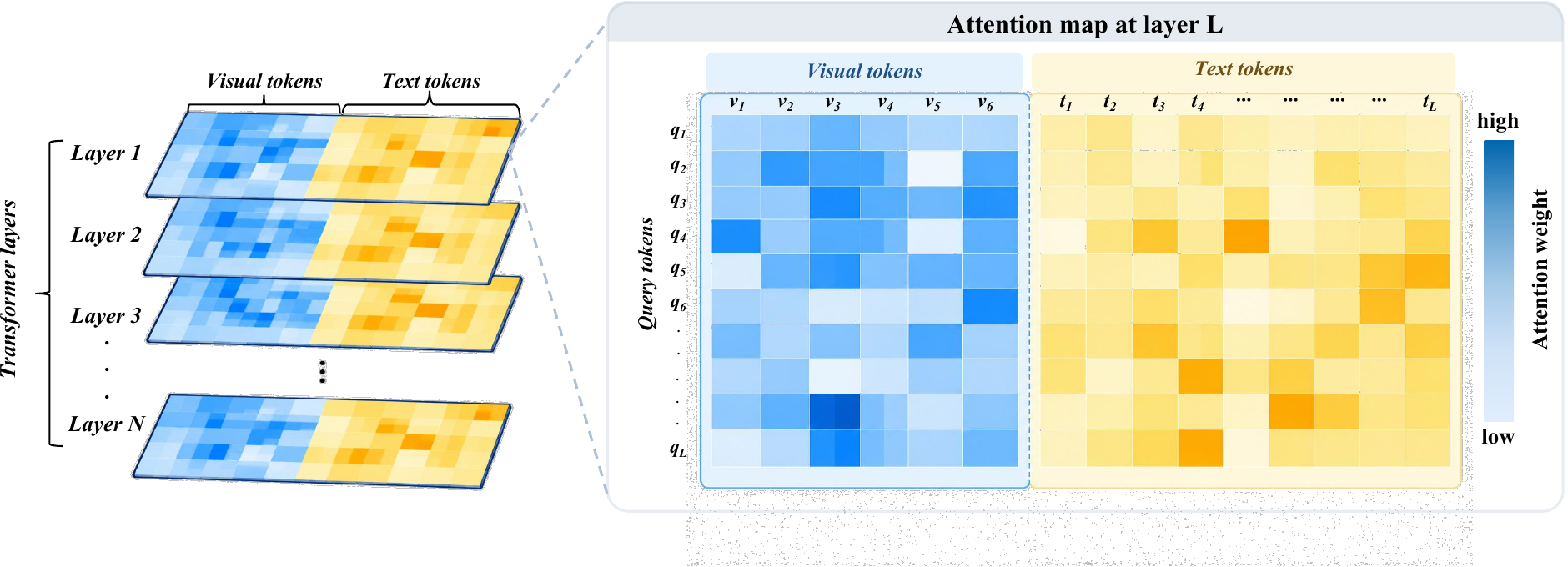}
\captionsetup{width=\textwidth}
\caption{Illustration of layer-wise visual attention used to compute visual reliance.}
\label{fig:visual-reliance}
\end{figure*}
\setcounter{figure}{3}

\subsubsection{Question-Likelihood and Visual Reliance}

For an observed question $q=(q_1,\ldots,q_L)$, let $o$ denote the questioner-visible non-image context and let $C$ denote the ordered candidate images. Under teacher forcing, we score $q$ by its conditional sequence likelihood:
\begin{equation}
p_\theta(q\mid o,C)=\prod_{\ell=1}^{L}p_\theta(q_\ell\mid q_{<\ell},o,C).
\end{equation}
For the likelihood comparisons below, we use its token-averaged negative log-likelihood in the IAS, CS, $\Delta_{\mathrm{R1}}$, and $\Delta_{\mathrm{Shuffle}}$ diagnostics:
\begin{equation}
\ell_{\mathrm{TF}}(q;o,C)=-\frac{1}{L}\sum_{\ell=1}^{L}\log p_\theta(q_\ell\mid q_{<\ell},o,C).
\end{equation}
The likelihood diagnostics are
\begin{equation}
\mathrm{IAS}(q)=\ell_{\mathrm{TF}}(q;o,\varnothing)-\ell_{\mathrm{TF}}(q;o,C).
\end{equation}
\begin{equation}
\mathrm{CS}(q)=\frac{1}{K}\sum_{k=1}^{K}\left[\ell_{\mathrm{TF}}(q;o,C\setminus c_k)-\ell_{\mathrm{TF}}(q;o,C)\right].
\end{equation}
\begin{equation}
\Delta_{\mathrm{R1}}=\ell_{\mathrm{TF}}(q;o,C_{\mathrm{R1}})-\ell_{\mathrm{TF}}(q;o,C).
\end{equation}
\begin{equation}
\Delta_{\mathrm{Shuffle}}=\ell_{\mathrm{TF}}(q;o,\pi(C))-\ell_{\mathrm{TF}}(q;o,C).
\end{equation}
Here $K=4$, $\varnothing$ denotes image ablation, $C\setminus c_k$ removes $c_k$, $C_{\mathrm{R1}}$ retains Rank-1, and $\pi(C)$ permutes Top-4. These four likelihood-based diagnostics measure image dependence (IAS) and sensitivity to candidate composition, Rank-1 retention, and order (CS, $\Delta_{\mathrm{R1}}$, and $\Delta_{\mathrm{Shuffle}}$, respectively). Positive values raise loss, absolute values measure sensitivity strength, and near-zero values indicate weak sensitivity. All differences use token-averaged $\ell_{\mathrm{TF}}$ and exclude prompt, dialogue, and image tokens.

We also examine visual reliance in candidate-conditioned question generation, following recent analyses of underused decisive regions in MLLM perception \citep{peng2026deeper,wei2026zooming,yuan2026visionopd}. Let $\alpha^{(n)}_{\ell,u}$ denote the attention weight averaged over all heads in layer $n$ from input position $u$ to output position $q_\ell$. Let $\mathcal V$ be the visual-token positions, $\mathcal U$ be the all attended input positions, and $N$ be the number of layers included. The Visual Reliance Ratio (VRR) is defined as follows; larger values indicate stronger visual conditioning:

\begin{equation}
\mathrm{VRR}(q)=\frac{\sum_{n=1}^{N}\sum_{\ell=1}^{L}\sum_{v\in\mathcal V}\alpha^{(n)}_{\ell,v}}
{\sum_{n=1}^{N}\sum_{\ell=1}^{L}\sum_{u\in\mathcal U}\alpha^{(n)}_{\ell,u}}.
\end{equation}

The corresponding layer-wise attention maps are shown in Figure~\ref{fig:visual-reliance}; darker cells indicate larger weights in the visual-token region.

\begin{figure}[t]
\centering
\includegraphics[width=\textwidth]{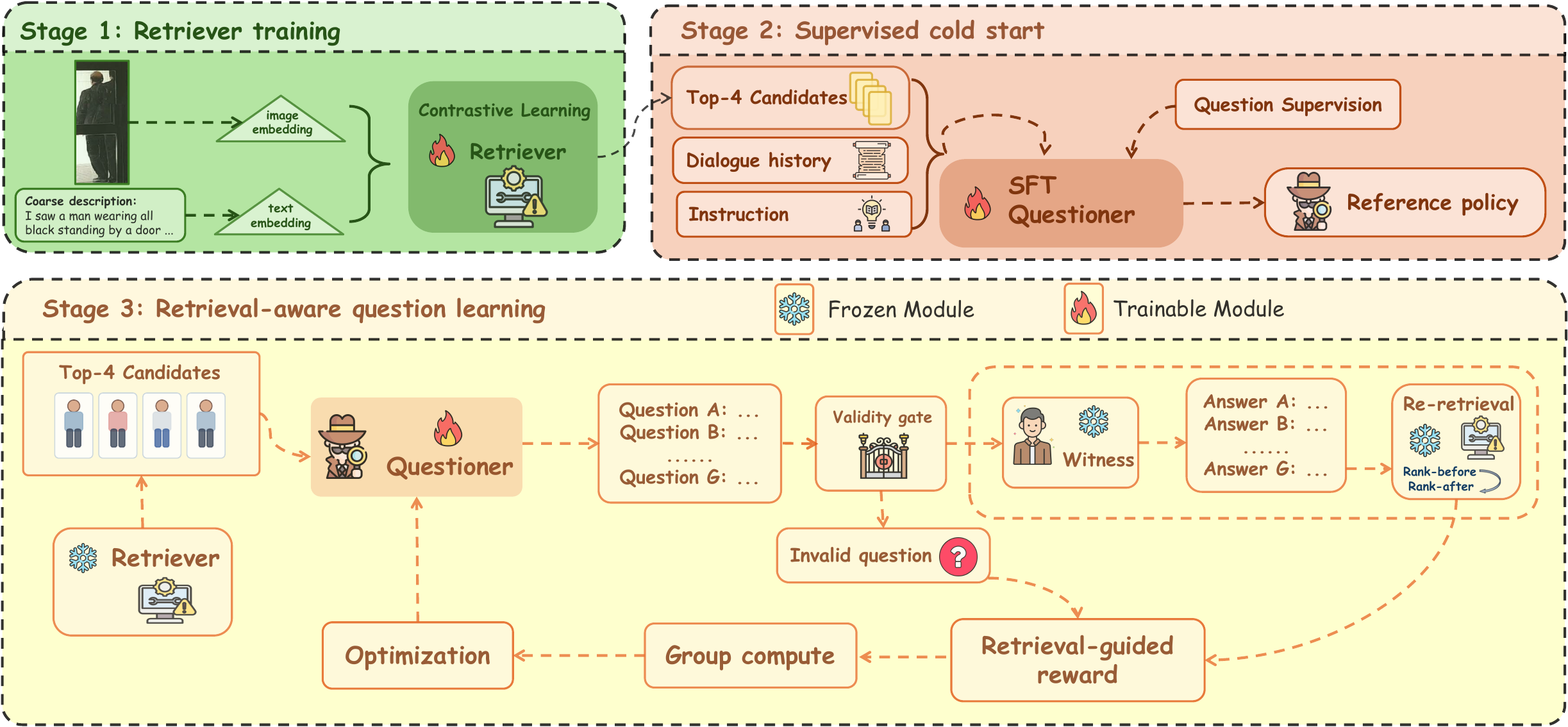}
\caption{Overview of RAVEL's three-stage training pipeline.}
\label{fig:ravel-pipeline}
\end{figure}
\setcounter{figure}{4}

Supplementary Material A reports the four likelihood diagnostics and VRR. IAS and VRR rise across rounds, whereas CS remains small and candidate retention barely changes question likelihood. These observations motivate optimizing questions by realized utility after answering and re-retrieval.

\subsection{Retriever}

We use CLIP-ViT-B/16 with the IRRA training framework \citep{jiang2023cross}, trained for 30 epochs on fine-grained Interactive-PEDES descriptions. The retriever is then frozen during questioner training and evaluation, providing a consistent ranking signal for the interaction loop. Full training settings are given in Supplementary Material B.

\subsection{Supervised Fine-Tuning Cold Start}

We initialize the questioner from LLaVA-OneVision-Qwen2-7B-ov \citep{li2024llavaonevision} and fine-tune it for one epoch with QLoRA \citep{dettmers2023qlora} on the official question-generation supervision. This supervised fine-tuning (SFT) checkpoint provides the starting policy for online optimization; SFT hyperparameters are listed in Supplementary Material B.

\subsection{Retrieval-Aware Reinforcement Learning}

\paragraph{Policy and partial observation.} RAVEL uses the supervised questioner from Section~3.3 as a cold start, then optimizes it online in the retrieval environment. At round $t$, the questioner samples an open-ended natural-language question from $\pi_\theta(q_t\mid o_t)$. Its partial observation $o_t$ contains the initial description, all previous questions and answers, the accumulated retrieval text, the ordered Top-4 candidates, and the round index. It excludes both the private witness memory $M$ and the target rank $r_t$: the witness memory is available only to the answerer, while the questioner must infer what to ask from the evolving dialogue and candidate context. The action space consists of free-form questions conditioned on the current candidate set and interaction history. For each training state, it samples a group of $G$ candidate questions from the same partial observation, so their realized outcomes can be compared under matched conditions.

\paragraph{Online question--answer--retrieval loop.} Each sampled question is executed in the same interaction environment: the fixed answerer receives the question together with the private witness memory, and its response is processed with the fixed answer-length limit and answer-cleaning rules. The cleaned answer is appended to the accumulated retrieval text, after which the frozen retriever scores the fixed gallery again and returns the target's new rank. This updated dialogue, retrieval text, candidate list, and round index form the next partial observation. The question is therefore evaluated through the evidence it actually elicits and the ranking change that evidence produces. The answerer and retriever remain fixed during reinforcement learning (RL), isolating learning to the question strategy. Because the policy is optimized against retrieval outcomes, it can also be paired with an alternative retrieval backbone without further questioner optimization.

\paragraph{Validity gate.} The gate is applied during both training and test-time rollouts to maintain the same answerable witness-interaction protocol across methods. It rejects candidate-selection or ranking requests, self-filled answers, memory-dump language, and overly broad prompts without a localized target. These outputs could inject information unavailable through a legitimate witness response or create a shortcut in the retrieval text. All methods retain the shared answer-cleaning protocol. Empty questions are excluded from the policy-loss update, while other invalid questions remain in group normalization and receive only the fixed invalid-action reward. The gate thus constrains protocol violations while retaining a free question vocabulary, leaving the policy free to explore answerable questions. The complete invalid-action taxonomy and trigger rules are given in Supplementary Material B.

\paragraph{Retrieval reward.} The reward uses realized reciprocal-rank improvement and gives additional credit for reaching Rank-1. Let $r_{\mathrm{before}}$ and $r_{\mathrm{after}}$ denote the target ranks before and after adding the processed answer. The reciprocal-rank component is
\begin{equation}
\Delta_{\mathrm{RR}}=\frac{1}{r_{\mathrm{after}}}-\frac{1}{r_{\mathrm{before}}}.
\end{equation}
With $\mathcal V$ denoting the set of questions that pass the validity gate, the reward is
\begin{equation}
R(q)=
\begin{cases}
\Delta_{\mathrm{RR}}+0.35\mathbf{1}[r_{\mathrm{before}}>1,\,r_{\mathrm{after}}=1]-0.02\mathbf{1}[\mathrm{group\text{-}repeat}], & q\in\mathcal V,\\[-1pt]
-0.2, & q\notin\mathcal V.
\end{cases}
\end{equation}
The $0.35$ Rank-1 entry bonus rewards reaching the top of the gallery when the target was not already Rank-1. The separate $\mathrm{group\text{-}repeat}$ term discourages duplicate samples within a GRPO group while leaving valid questions subject to normal interaction evaluation. We do not directly reward a predefined question category, question length, or explicit visual terminology. The objective therefore lets the policy discover useful evidence through the answer-and-retrieval outcome across states and rounds. The main experiment uses the curated 3K-state configuration; state sampling and reward coefficients are detailed in Supplementary Material B.

\paragraph{Group-relative policy update.} Starting from the SFT checkpoint, we optimize the questioner with Group Relative Policy Optimization (GRPO) \citep{shao2024deepseekmath}. For each state, the policy samples $G$ questions, evaluates each through the interaction loop above, and normalizes their rewards within the group to obtain relative advantages. The resulting advantages drive a token-level clipped policy objective with an importance ratio and a K3 reference-policy penalty; the SFT checkpoint is the reference, and the loss is applied only to generated question tokens. The group-normalized advantage, importance ratio, reference penalty, and clipped objective are specified in Supplementary Material B. We update only the questioner's QLoRA adapter, keeping the vision tower and retriever frozen while leaving the answerer unchanged. The selector is not used during RAVEL training or inference: the policy receives the current Top-4 candidates directly. The primary policy is trained on the curated 3K-state pool, with group construction, optimizer settings, and decoding parameters reported in Supplementary Material B.

\section{Experiments}

\subsection{Experimental Setup}

We use CLIP \citep{radford2021clip} with IRRA \citep{jiang2023cross} as the frozen retriever, initialize the questioner from LLaVA-OneVision-Qwen2-7B-ov with QLoRA \citep{dettmers2023qlora}, and use Qwen2.5-7B-Instruct \citep{yang2024qwen25} as the answerer in the main Interactive-PEDES experiments. The answerer receives the fine-grained witness description and the generated question. RAVEL directly receives the current Top-4 results and learns from 3,000 training states sampled across rounds and source datasets. We evaluate five interaction rounds on the 7,373-query Interactive-PEDES test split and report Rank-1, Rank-5, Rank-10, mAP, and Best log Rank Integral (BRI) \citep{lee2024plugir}. BRI summarizes retrieval quality over the interaction trajectory, with lower values indicating better cumulative retrieval. All main-comparison methods share the same answer and text-cleaning protocol.

\subsection{Interactive Retrieval and Questioning Behavior}

\setcounter{table}{0}
\begin{table*}[t]
\centering
\caption{Interactive retrieval performance on Interactive-PEDES.}
\label{tab:main-interactive}
\resizebox{\textwidth}{!}{%
\begin{tabular}{l|cccc|cccc|c}
\toprule
\multirow{2}{*}{\centering Method} & \multicolumn{4}{c|}{Round 3} & \multicolumn{4}{c|}{Round 5} & \multirow{2}{*}{\centering BRI $\downarrow$} \\
& $R@1$ & $R@5$ & $R@10$ & mAP & $R@1$ & $R@5$ & $R@10$ & mAP & \\
\midrule
Initial & 37.61 & 61.74 & 72.01 & 28.23 & 37.61 & 61.74 & 72.01 & 28.23 & - \\
PlugIR \citep{lee2024plugir} & 43.15 & 67.17 & 76.78 & 30.92 & 47.15 & 70.83 & 79.73 & 33.80 & 1.012 \\
ChatIR \citep{levy2023chatting} & 45.02 & 69.24 & 78.60 & 33.07 & 49.06 & 72.94 & 81.57 & 35.85 & 0.997 \\
GPT-5.6 Luna \citep{openai2026gpt56luna} & 46.37 & 67.73 & 77.54 & 35.43 & 47.88 & 69.27 & 78.19 & 36.56 & 1.042 \\
SimRV \citep{liang2023simple} & \underline{62.91} & \underline{83.26} & \underline{89.18} & 38.65 & 63.42 & 84.21 & 89.34 & 39.46 & 0.720 \\
LLaVA-ReID \citep{lu2025llavareid} & 60.11 & 81.41 & 88.31 & \underline{40.49} & \underline{67.94} & \underline{86.82} & \underline{92.47} & \underline{45.16} & \underline{0.703} \\
\midrule
RAVEL (ours) & \textbf{64.13} & \textbf{84.28} & \textbf{90.40} & \textbf{42.81} & \textbf{73.73} & \textbf{90.53} & \textbf{94.98} & \textbf{47.89} & \textbf{0.642} \\
\bottomrule
\end{tabular}}
\end{table*}

We compare RAVEL with PlugIR \citep{lee2024plugir}, ChatIR \citep{levy2023chatting}, SimRV \citep{liang2023simple}, LLaVA-ReID \citep{lu2025llavareid}, GPT-5.6 Luna \citep{openai2026gpt56luna}, and the no-interaction Initial setting under the same evaluation protocol. Table~\ref{tab:main-interactive} shows that RAVEL achieves the strongest retrieval results at both evaluation rounds; underlined entries mark the second-best result in each column, and a dash denotes an unavailable or inapplicable metric. At Round 5, it reaches 73.73 R@1 and reduces BRI to 0.642, improving the standard selector-based LLaVA-ReID pipeline by 5.79 Rank-1 points. The matched input and retriever comparisons are reported in Table~\ref{tab:input-retriever-ablation}. RAVEL improves over Initial as dialogue accumulates. The following analysis examines the question allocation and retrieval evidence learned by the policy.

\paragraph{Closed-source vision-language model.} GPT-5.6 Luna \citep{openai2026gpt56luna} follows the same candidate, answerer, text-cleaning, and evaluation protocol. After five rounds, it reaches 47.88 Rank-1, 36.56 mAP, and 1.042 BRI; prompt and decoding details are provided in Supplementary Material B.

\paragraph{Questioning behavior and retrieval utility.}

\begin{figure}[t]
\centering
\includegraphics[width=\textwidth]{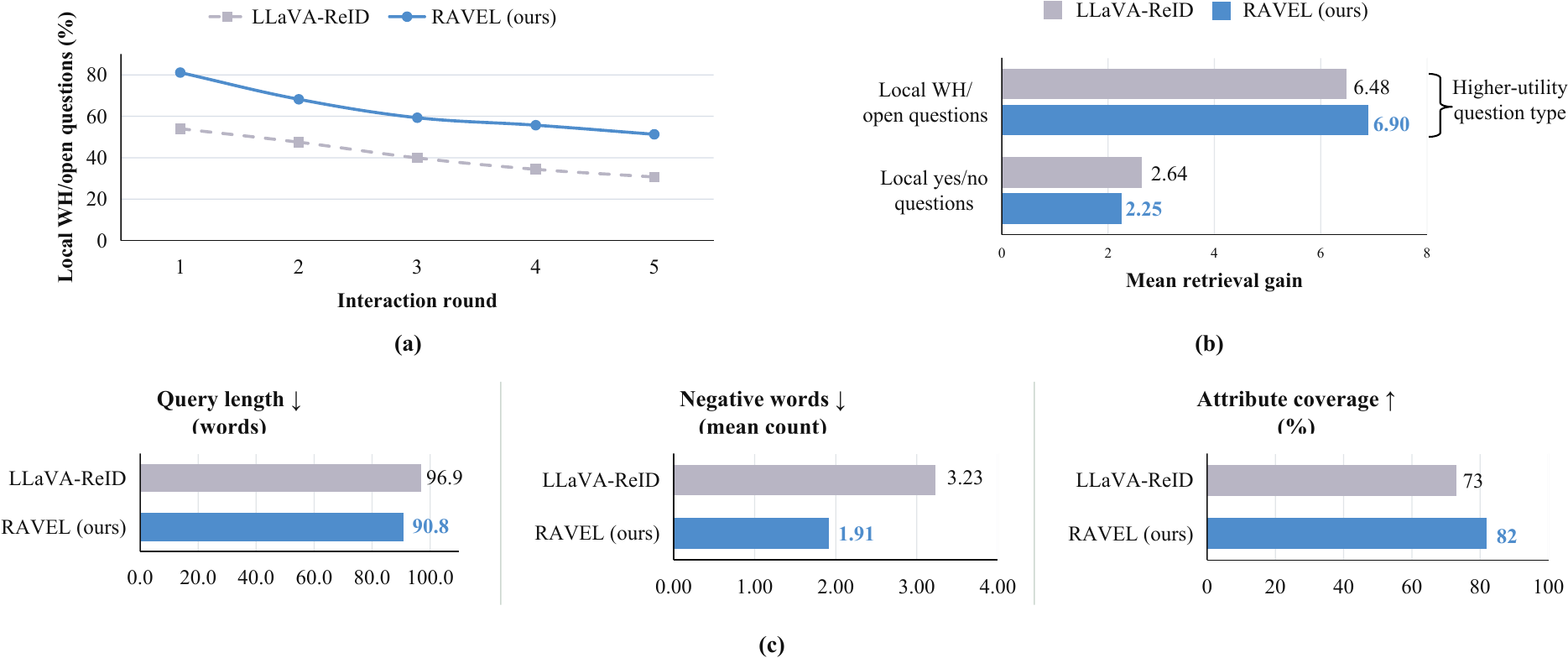}
\captionsetup{width=\textwidth}
\caption{Analysis of learned questioning behavior and retrieval-text quality. (a) Ratio of local WH/open questions over interaction rounds. (b) Mean retrieval gain of different question types. (c) Quality of the final retrieval text, measured by query length, negative-word count, and attribute coverage.}
\label{fig:behavior-analysis}
\end{figure}

RAVEL improves retrieval by learning a state-dependent allocation of the five question turns toward localized attributes, especially hair and head cues (Supplementary Figure~6). In Figure~\ref{fig:behavior-analysis}, the question-type distribution in (a) and the retrieval-gain comparison in (b) jointly explain the improvement: the share of local WH/open questions declines across rounds for both methods, but remains consistently higher for RAVEL, ending at 54\% versus 31\% for LLaVA-ReID. Local WH/open questions also yield larger mean retrieval gains than local yes/no questions: 6.90 vs. 2.25 for RAVEL and 6.48 vs. 2.64 for LLaVA-ReID. RAVEL therefore allocates more turns to this stronger-evidence question type, then uses targeted verification to refine the evidence it elicits.

\clearpage

Beyond question allocation, Figure~\ref{fig:behavior-analysis}(c) summarizes how the learned policy shapes the final retrieval text. The resulting retrieval text becomes shorter while retaining more useful evidence. Across all 7,373 queries in the Interactive-PEDES test set, the five-round description decreases from 96.9 to 90.8 words, while the mean negative-word count drops from 3.23 to 1.91. Attribute coverage increases from 73\% to 82\%; the largest additions concern shoes (+10.8 points), environment (+9.6), lower-body clothing (+5.5), and hair (+4.6). Fewer negative and other noisy words reduce interference and concentrate the retrieval text on positive, visually grounded attributes, helping explain how RAVEL converts the interaction budget into more effective evidence.

\subsection{Transfer to Text-based ReID}

\setcounter{table}{1}
\begin{table*}[t]
\centering
\caption{Comparison of text-based ReID retrievers and interactive questioners on three benchmarks.}
\label{tab:transfer}
\resizebox{\textwidth}{!}{%
\begin{tabular}{l|cccc|cccc|cccc}
\toprule
\multirow{2}{*}{\centering Method} & \multicolumn{4}{c|}{\textbf{CUHK-PEDES}} & \multicolumn{4}{c|}{\textbf{ICFG-PEDES}} & \multicolumn{4}{c}{\textbf{RSTPReid}} \\
& $R@1$ & $R@5$ & $R@10$ & mAP & $R@1$ & $R@5$ & $R@10$ & mAP & $R@1$ & $R@5$ & $R@10$ & mAP \\
\midrule
CFine \citep{yan2023cfine} & 69.57 & 85.93 & 91.15 & - & 60.83 & 76.55 & 82.42 & - & 50.55 & 72.50 & 81.60 & - \\
RaSa \citep{bai2023rasa} & 76.51 & 76.51 & 94.25 & 69.38 & 65.28 & 80.40 & 85.12 & 41.29 & 66.90 & 86.50 & 91.35 & 52.31 \\
APTM \citep{yang2023aptm} & 76.53 & 90.04 & 94.15 & 66.91 & 68.51 & 82.99 & 87.56 & 41.22 & 67.50 & 85.70 & 91.45 & 52.56 \\
AUL \citep{li2024aul} & 77.23 & 90.43 & 94.41 & - & 69.16 & 83.32 & 88.37 & - & 71.65 & 87.55 & 92.05 & - \\
RDE \citep{qin2024rde} & 76.20 & 90.53 & 94.20 & 67.85 & 67.83 & 82.50 & 87.28 & 40.79 & 67.40 & 85.65 & 90.60 & 52.01 \\
IRRA \citep{jiang2023cross} & 73.44 & 89.36 & 93.34 & 66.09 & 63.57 & 80.36 & 85.78 & 38.17 & 59.30 & 81.50 & 88.50 & 47.69 \\
\midrule
IRRA \citep{jiang2023cross} + LLaVA-ReID \citep{lu2025llavareid} & 78.51 & 92.43 & 95.67 & 70.61 & 67.44 & 82.91 & 87.69 & 40.60 & 69.85 & 88.10 & 92.55 & 54.92 \\
IRRA \citep{jiang2023cross} + RAVEL & 80.65 & 94.35 & 97.86 & 72.49 & 69.51 & 84.75 & 89.80 & 42.55 & 72.03 & 90.01 & 94.68 & 56.85 \\
\midrule
\bottomrule
\end{tabular}}

\end{table*}

We integrate RAVEL with existing text-based ReID frameworks and evaluate transferability on CUHK-PEDES \citep{li2017cuhkpedes}, ICFG-PEDES \citep{ding2021ssan}, and RSTPReid \citep{zhu2021dssl}. The comparison covers CFine \citep{yan2023cfine}, RaSa \citep{bai2023rasa}, APTM \citep{yang2023aptm}, and AUL \citep{li2024aul}. The scoring protocol is detailed in Supplementary Material B. Dataset captions provide the initial retrieval text, and the trained questioner and IRRA retriever are reused without retraining for five interaction rounds. The base T-ReID model encodes the initial caption, while the interactive retriever encodes the accumulated dialogue; their matching scores are averaged for final re-ranking.

Table~\ref{tab:transfer} reports the cross-dataset transfer results, placing RDE \citep{qin2024rde} and IRRA \citep{jiang2023cross} alongside other conventional text-based ReID methods and comparing IRRA \citep{jiang2023cross} with its interactive counterparts; underlined entries mark the second-best result in each column, and a dash denotes an unavailable or inapplicable metric. The RDE \citep{qin2024rde} retriever substitution is evaluated separately in Table~\ref{tab:input-retriever-ablation}. Using IRRA \citep{jiang2023cross} as the retriever, RAVEL raises Rank-1 by 7.21 points on CUHK-PEDES, 5.94 points on ICFG-PEDES, and 12.73 points on RSTPReid. Its mAP gains over the IRRA \citep{jiang2023cross} backbone are 6.40, 4.38, and 9.16 points, respectively. The consistent gains across three benchmarks indicate that retrieval-aware question learning complements standard cross-modal retrieval models beyond Interactive-PEDES.

\subsection{Ablation Study}

\setcounter{table}{2}
\noindent
\begin{minipage}[t]{0.48\textwidth}
\centering
\captionof{table}{Ablation of the training procedure.}
\label{tab:train-ablation}
\fontsize{7pt}{8.4pt}\selectfont
\setlength{\tabcolsep}{2pt}
\renewcommand{\arraystretch}{1.05}
\begin{tabular}{lccc|rrrrr}
\toprule
Model & SFT & RL & Gate & R@1 & R@5 & R@10 & mAP & BRI \\
\midrule
LLaVA-OV & & & & 44.12 & 68.24 & 77.73 & 32.55 & 1.075 \\
LLaVA-OV & \checkmark & & & 69.44 & 87.78 & 92.80 & 45.48 & 0.698 \\
LLaVA-OV & & \checkmark & \checkmark & 65.66 & 84.66 & 90.21 & 40.30 & 0.698 \\
LLaVA-OV & \checkmark & \checkmark & & \multicolumn{5}{c}{\textit{Collapse}} \\
LLaVA-OV & \checkmark & \checkmark & \checkmark & 73.73 & 90.53 & 94.98 & 47.89 & 0.642 \\
\bottomrule
\end{tabular}
\end{minipage}

\hfill
\begin{minipage}[t]{0.48\textwidth}
\centering
\captionof{table}{Ablation of candidate input and retriever choice after five interaction rounds.}
\label{tab:input-retriever-ablation}
\scriptsize
\setlength{\tabcolsep}{3pt}
\resizebox{\linewidth}{!}{%
\begin{tabular}{lrrrrr}
\toprule
Configuration & R@1 & R@5 & R@10 & mAP & BRI $\downarrow$ \\
\midrule
\multicolumn{6}{l}{\textit{Candidate input}} \\
LLaVA-ReID (selector) & 67.94 & 86.82 & 92.47 & 45.16 & 0.703 \\
LLaVA-ReID (direct Top-4) & 67.84 & 87.54 & 92.72 & 45.33 & 0.701 \\
RAVEL (direct Top-4) & \textbf{73.73} & \textbf{90.53} & \textbf{94.98} & \textbf{47.89} & \textbf{0.642} \\
\midrule
\multicolumn{6}{l}{\textit{Retriever}} \\
RAVEL + IRRA \citep{jiang2023cross} & 73.73 & 90.53 & 94.98 & 47.89 & 0.642 \\
RAVEL + RDE \citep{qin2024rde} & \textbf{77.53} & \textbf{92.08} & \textbf{95.56} & \textbf{49.06} & \textbf{0.614} \\
\bottomrule
\end{tabular}
}
\end{minipage}

\par

We study the training recipe and input/retriever configuration under the same four-candidate, answer-length, and cleaning settings. The RL state-pool scaling study is reported in Supplementary Material B.

\paragraph{Training-procedure ablation.}

Table~\ref{tab:train-ablation} separates the native questioner, SFT-only, RL-only, and SFT-to-RL paths. Rank-1 rises from 44.12 for the native model to 69.44 with SFT, 65.66 with RL from the native initialization, and 73.73 with SFT followed by gated RL. Removing the gate causes a rapid invalid-action collapse: the policy starts emitting candidate-person references that prompt broad ``candidate person'' descriptions, together with too-short or non-question outputs, and these patterns dominate the rollouts as training proceeds. This behavior exposes a rank-only reward loophole, whereas the gate keeps the policy focused on answerable questions.

\paragraph{Candidate input and retriever ablation.}

Table~\ref{tab:input-retriever-ablation} isolates candidate access and retriever choice within the Interactive-PEDES protocol. The RDE \citep{qin2024rde} row replaces frozen IRRA \citep{jiang2023cross} after RAVEL training, while the questioner and answerer remain unchanged. Direct Top-4 input alone changes little for LLaVA-ReID, whereas pairing it with RAVEL's online learning yields stronger gains, showing that the improvement comes from learning retrieval-useful questions. Replacing IRRA with RDE after training further improves retrieval without updating RAVEL, indicating transferability across retriever backbones.

\FloatBarrier

\subsection{Qualitative Analysis}

We select two cases by baseline final rank: one near the gallery front and one recovery case. Supplementary Material C gives the five-round records. RAVEL replaces repeated verification with localized attribute questions and improves the target rank through sequential evidence acquisition.
At inference, RAVEL receives no ground-truth rank, target image, or oracle attribute list; the cases show how question timing and local attribute choice shape the final ranking.

\section{Conclusion}

RAVEL frames interactive person ReID as sequential evidence acquisition. It starts from supervised question generation, observes the current Top-4 candidates and dialogue state, and learns from rank feedback after each answer updates the retrieval text. The diagnostic study shows that static question judgments and candidate-set perturbations are weak proxies for downstream retrieval utility, motivating online optimization in the realized interaction loop. Behavioral analysis further shows that the learned policy favors localized open-ended attribute questions, which yield stronger retrieval gains and more focused evidence. The ablations confirm the importance of the validity gate and show that the learned question strategy remains compatible with alternative retrievers. Together, these findings support retrieval-aware question learning as a practical way to make interactive ReID more adaptive and evidence-driven.

\clearpage
\section*{AI Use Statement}

In this work, we have not used generative AI tools for any task with required disclosure, and the remaining required-disclosure tasks are not applicable to this work. Additionally, we used generative AI tools to draft sections of the paper and to aid or polish the writing. We have reviewed all AI-assisted work: the authors verified the final wording, claims, equations, experimental settings, tables, figures, and reported results against the project records. We take responsibility for the final content of this work, including text, claims, and artifacts produced with the aid of generative AI.

\section*{Ethics Statement}

This work uses publicly available person re-identification benchmarks and follows their stated terms of use and original dataset documentation. It does not recruit participants, conduct interventions, or collect personally identifying information. The human-labeled subset contains task-specific judgments on benchmark examples only; annotators were not asked to provide personal or sensitive information. We do not attempt to identify individuals or infer sensitive attributes, and the qualitative examples are used solely to illustrate retrieval behavior. Because benchmark images and annotations may reflect biases in their collection and labeling, the reported results should be interpreted as benchmark retrieval performance; they do not establish fairness in deployment.

\section*{Reproducibility Statement}

The main paper and supplementary material document the information needed to reproduce the study. Section 3 describes the diagnostic analyses, RAVEL method, environment, reward, and optimization procedure, while Section 4 specifies the datasets, evaluation protocol, baselines, and ablations. Supplementary Material A defines the diagnostic measures and label mappings; Supplementary Material B records the implementation settings, prompts, decoding configuration, invalid-action rules, and compute details; and Supplementary Material C provides representative interaction trajectories. The code, training configurations, evaluation scripts, and processed interaction logs used in this work will be publicly released upon publication. We use publicly available benchmarks and will provide the corresponding data-processing scripts, split information, and evaluation details; the original datasets will not be redistributed. Together, these materials provide the methodological, experimental, and qualitative references needed to reproduce the reported results.

\bibliographystyle{iclr2027_conference}
\bibliography{references}

\clearpage
\setcounter{page}{14}
\iclrrulercount=648
\setcounter{table}{4}
\setcounter{figure}{5}
\setcounter{equation}{9}

\section*{A. Question-Value Diagnostics}

Sections 3.1.1--3.1.2 of the main paper define the four likelihood diagnostics and VRR. This section records the sampling protocol, statistical estimators, and annotation details used to reproduce the analysis.

\begin{table}[H]
\centering
\begin{minipage}[t]{0.48\textwidth}
\centering
\captionsetup{width=\linewidth}
\captionof{table}{Likelihood diagnostics and VRR of the LLaVA-ReID questioner on Interactive-PEDES.}
\label{tab:visual-reliance}
\scriptsize
\setlength{\tabcolsep}{1.5pt}
\resizebox{\linewidth}{!}{%
\begin{tabular}{c|rrrrr}
\toprule
Round & IAS $\uparrow$ & CS $\uparrow$ & $\Delta_{\mathrm{R1}}$ & $\Delta_{\mathrm{Shuffle}}$ & VRR $\uparrow$ \\
\midrule
0 & 0.0272 & 0.0042 & -0.0000 & -0.0115 & 0.0631 \\
1 & 0.0369 & 0.0066 &  0.0001 & -0.0168 & 0.0869 \\
2 & 0.0460 & 0.0080 & -0.0011 & -0.0199 & 0.0945 \\
3 & 0.0506 & 0.0090 & -0.0004 & -0.0247 & 0.0965 \\
4 & 0.0542 & 0.0091 &  0.0002 & -0.0251 & 0.0970 \\
\bottomrule
\end{tabular}}
\end{minipage}
\hfill
\begin{minipage}[t]{0.48\textwidth}
\centering
\captionsetup{width=\linewidth}
\captionof{table}{Per-round question-quality statistics on the 2,000-state model-annotated sample.}
\label{tab:diagnostic-summary}
\begingroup
\scriptsize
\setlength{\tabcolsep}{2pt}
\resizebox{\linewidth}{!}{%
\begin{tabular}{c|r|r|r|r|r|r|r|r}
\toprule
Round & $n$ & Improve & $d+$ & $u+$ & $\rho_d$ & $\rho_u$ & $r_d$ & $r_u$ \\
\midrule
0 & 400 & 25.00\% & 41.00\% & 39.00\% & 0.034 & 0.008 & 0.082 & 0.047 \\
1 & 400 & 25.00\% & 38.00\% & 37.75\% & 0.017 & 0.012 & 0.059 & 0.051 \\
2 & 400 & 25.00\% & 34.50\% & 34.25\% & 0.088 & 0.080 & 0.140 & 0.131 \\
3 & 400 & 25.00\% & 32.75\% & 32.25\% & 0.080 & 0.087 & 0.151 & 0.157 \\
4 & 400 & 25.00\% & 30.50\% & 30.00\% & 0.083 & 0.094 & 0.107 & 0.113 \\
\bottomrule
\end{tabular}
}%
\endgroup
\end{minipage}
\end{table}

The offline-supervision diagnostic contains 2,000 interaction states sampled from observed test-set rollouts. The sample is balanced across the five interaction rounds (400 states per round) and stratified by outcome: 500 rank-improvement states and 1,500 non-improvement states. Each state stores the current Top-4 candidates, dialogue history, question, answer, and rank change. Qwen3-VL-32B receives the state and returns structured judgments for candidate discrimination, shared attributes, absent or invisible attributes, generic background content, negative or unknown answers, redundancy with history, visual uncertainty, and likely retrieval usefulness. We use 101 human-annotated cases for the human--model agreement analysis.

For each state $i$, let $d_i\in\{0,1\}$ denote the model's discriminative-attribute judgment, $u_i\in\{0,1\}$ its likely-useful judgment, $\Delta_{\mathrm{RR},i}=1/r_i^{\mathrm{after}}-1/r_i^{\mathrm{before}}$ the reciprocal-rank gain, and $z_i=\mathbf{1}[\Delta_{\mathrm{RR},i}>0]$ the rank-improvement indicator. We compute Spearman's rank correlation between each binary judgment and $\Delta_{\mathrm{RR}}$ by applying average ranks to tied values:

\begin{equation}
\rho_s(x,\Delta_{\mathrm{RR}})=\operatorname{Corr}\!\left(\operatorname{rank}(x),\operatorname{rank}(\Delta_{\mathrm{RR}})\right).
\end{equation}
For the binary improvement event, we use the point-biserial correlation:
\begin{equation}
r_{\mathrm{pb}}(x,z)=\frac{\bar{x}_1-\bar{x}_0}{s_x}\sqrt{\frac{n_1n_0}{n^2}},
\end{equation}
where $\bar{x}_1$ and $\bar{x}_0$ are the means of $x$ among improved and non-improved states, $s_x$ is the sample standard deviation, and $n_1,n_0$ are the corresponding counts. For the human labels, "useless," "partial," and "useful" are mapped to $0$, $0.5$, and $1$ for continuous analyses. For agreement, "partial" and "useful" are grouped as a positive label. With $h_i$ and $m_i$ denoting the resulting human and model binary labels, Cohen's kappa is
\begin{equation}
\kappa=\frac{p_o-p_e}{1-p_e}, \qquad p_o=\frac{1}{n}\sum_i\mathbf{1}[h_i=m_i],
\end{equation}
where $p_e$ is the chance agreement obtained from the two annotators' marginal positive and negative rates. On the 2,000-state sample, the overall values are $\rho_s(d,\Delta_{\mathrm{RR}})=0.051$, $\rho_s(u,\Delta_{\mathrm{RR}})=0.048$, $r_{\mathrm{pb}}(d,z)=0.107$, and $r_{\mathrm{pb}}(u,z)=0.099$. For the 101 human-annotated cases, mapping ``useless,'' ``partial,'' and ``useful'' to $0$, $0.5$, and $1$ gives $\rho_s(h,\Delta_{\mathrm{RR}})=-0.016$ ($p=0.875$); grouping ``partial'' and ``useful'' as positive gives $r_{\mathrm{pb}}(h,z)=-0.041$ ($p=0.687$). The paired human--model audit gives $p_o=0.654$ and $\kappa=0.312$ for 101 completed labels.

\begin{table}[H]
\centering
\caption{Prompt used by the Qwen3-VL question-quality evaluator.}
\label{tab:prompt-quality-evaluator}
\includegraphics[width=\textwidth]{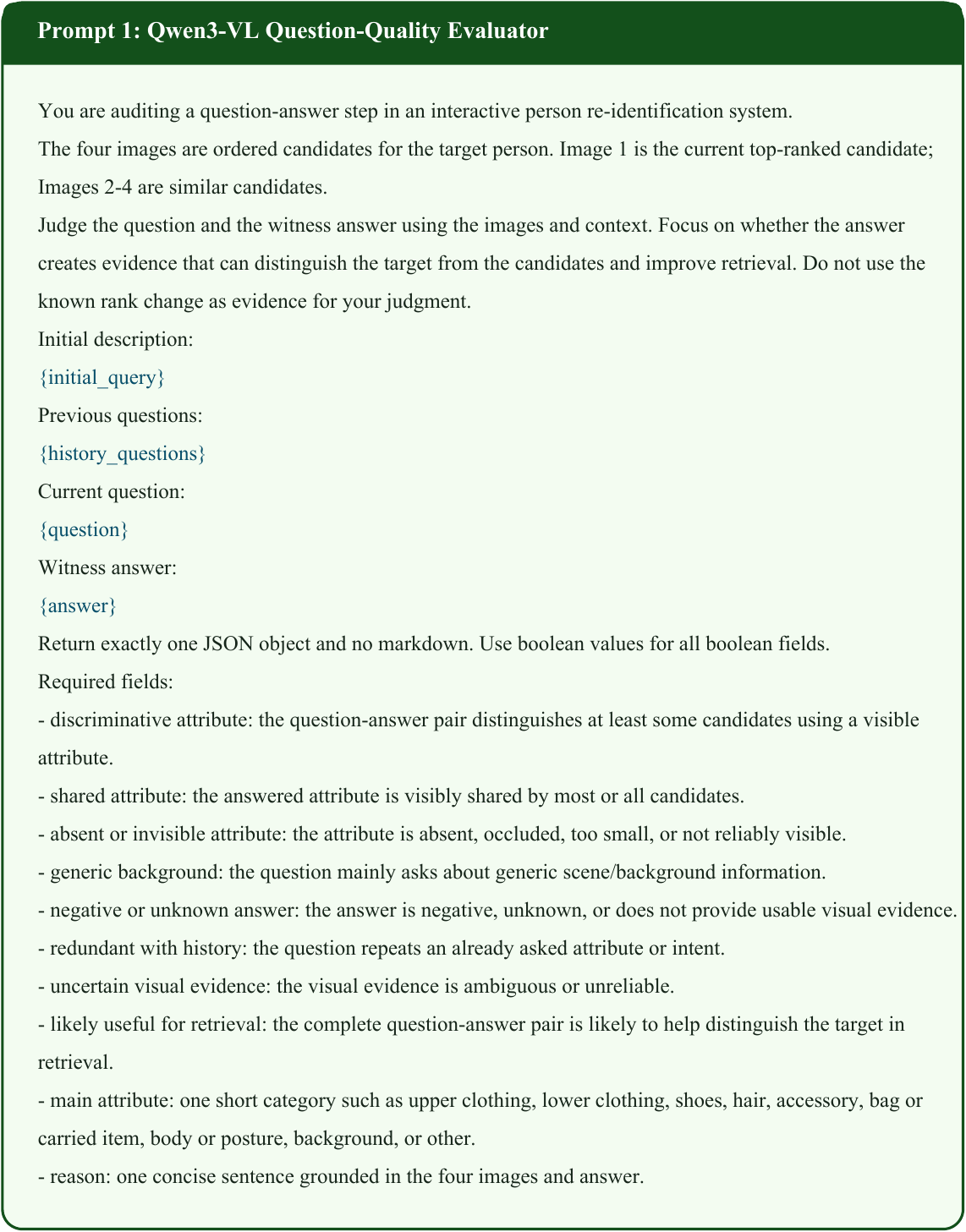}
\end{table}

\clearpage
\section*{B. Implementation Details}

\paragraph{Retriever.} We use CLIP-ViT-B/16 with the IRRA training framework. The retriever is trained for 30 epochs on fine-grained Interactive-PEDES descriptions with batch size 128, and is frozen for questioner training and evaluation. The text encoder uses the extended positional-embedding setting required by the long descriptions.

\paragraph{Questioner.} The questioner is initialized from LLaVA-OneVision-Qwen2-7B-ov. Supervised fine-tuning (SFT) uses QLoRA with rank 128, alpha 256, dropout 0.05, 4-bit NF4 quantization with double quantization, bf16 computation, learning rate \(1\times10^{-5}\), batch size 4, gradient accumulation 4, one epoch, cosine scheduling, 2\% warm-up, weight decay 0, and a maximum sequence length of 4096. The question length is limited to 96 tokens during interaction.

\paragraph{Answerer and interaction.} The main Interactive-PEDES experiments use Qwen2.5-7B-Instruct; it receives the fine-grained witness description and the generated question, with an answer length limit of 64 tokens. The shared answer-cleaning rule and fixed five-round interaction budget are retained. The LLaVA-ReID baseline uses its original selector, whereas RAVEL directly passes the current Top-4 candidates to the questioner and does not use a selector. The original dataset caption initializes retrieval, while the trained questioner and IRRA retriever are reused without retraining; after five rounds, the base and interactive retriever similarities are averaged for final ranking.

\paragraph{Reinforcement learning (RL) state construction.} The Interactive-PEDES training split contains 47,376 images and 11,543 identities. We construct the RL state pool from existing multi-round interaction records, sampling only training states whose target is not already Rank-1. The pool is stratified across interaction rounds and preserves the original proportions of the two source datasets, so that the policy observes different dialogue-history lengths and retrieval stages without introducing split or source imbalance. The primary experiment uses a curated 3,000-state pool; 1K and 5K pools use the same construction for controlled scaling analysis.

\begin{table}[H]
\centering
\begin{minipage}[t]{0.48\textwidth}
\centering
\captionsetup{width=\linewidth}
\captionof{table}{Primary-run scale, action validity, and compute cost.}
\label{tab:supp-cost}
\scriptsize
\renewcommand{\arraystretch}{0.88}
\resizebox{\linewidth}{!}{%
\begin{tabular}{ll}
\toprule
\textbf{Category} & \textbf{Value} \\
\midrule
RL training states & 3,000 \\
Question rollouts & 24,000 (8 per state) \\
Optimizer updates & 1,500 \\
Invalid-gate trigger rate & 2.02\% \\
Valid-action rate & 97.98\% \\
Mean unique questions / group & 7.93 / 8 \\
Mean repetition rate & 9.20\% \\
\midrule
Training hardware & 2 $\times$ NVIDIA A800 \\
Training wall-clock time & 13 h 15 min \\
Training compute & 26.5 GPU-hours \\
Five-round evaluation time & 8 h 19 min \\
Evaluation compute & 16.6 GPU-hours \\
\bottomrule
\end{tabular}
}%
\end{minipage}
\hfill
\begin{minipage}[t]{0.48\textwidth}
\centering
\captionsetup{width=\linewidth}
\captionof{table}{Scaling study of the curated RL state pool.}
\label{tab:data-ablation}
\scriptsize
\setlength{\tabcolsep}{3pt}
\renewcommand{\arraystretch}{1.05}
\resizebox{\linewidth}{!}{%
\begin{tabular}{lrrrrr}
\toprule
RL states & R@1 & R@5 & R@10 & mAP & BRI \\
\midrule
0K (SFT-only) & 69.44 & 87.78 & 92.80 & 45.48 & 0.698 \\
1K & 69.46 & 87.97 & 93.03 & 45.59 & 0.689 \\
3K & 73.73 & 90.53 & 94.98 & 47.89 & 0.642 \\
5K & 74.35 & 90.71 & 94.89 & 48.09 & 0.635 \\
\bottomrule
\end{tabular}
}%
\end{minipage}
\end{table}

\paragraph{RL optimization and cost.} RAVEL starts from the SFT checkpoint and updates only the QLoRA adapter. The primary study uses the curated 3,000-state pool, group size $G=8$, token-level clipped group-relative policy updates, the SFT checkpoint as the reference policy, and the K3 token-level KL estimator. We use AdamW with learning rate $1\times10^{-6}$, constant scheduling (no warm-up or decay), zero weight decay, and gradient-norm clipping at 1.0. Gradients accumulate over two state-level groups, so each optimizer update uses 16 sampled questions from two states; one epoch over 3,000 states therefore gives 1,500 optimizer updates and 24,000 question rollouts. During RL rollouts, generation uses sampling with temperature 1.0, top-$p=0.95$, no top-$k$ or beam search, and at most 100 new tokens. During test-time RAVEL evaluation, sampling uses temperature 1.0 (the model default), top-$p=0.5$, no top-$k$ or beam search, and at most 100 new tokens. During both training and inference, the policy receives the current Top-4 candidates directly. The complete reward, objective, and state-construction details are given below, while invalid-action rules are listed in the following subsection.

The main paper defines the reciprocal-rank reward and validity-gated reward function. For completeness, the remaining policy quantities are specified below.
For completeness, for a sampled group with rewards $r_1,\ldots,r_G$, we use the normalized advantage
\begin{equation}
A_i=\frac{r_i-\mu_G}{\max(\sigma_G,\varepsilon_{\mathrm{adv}})}.
\end{equation}
Here $\varepsilon_{\mathrm{adv}}=10^{-6}$ is the advantage-normalization floor.
For generated token $y_{i,t}$, the importance ratio is
\begin{equation}
\rho_{i,t}=\exp\!\left(\log\pi_\theta(y_{i,t}\mid x_i,y_{i,<t})-\log\pi_{\mathrm{old}}(y_{i,t}\mid x_i,y_{i,<t})\right),
\end{equation}
and the K3 reference penalty is
\begin{equation}
d^{\mathrm{K3}}_{i,t}=\exp(\delta_{i,t})-\delta_{i,t}-1,\qquad
\delta_{i,t}=\log\pi_{\mathrm{ref}}-\log\pi_\theta.
\end{equation}
The optimized objective is
\begin{equation}
\mathcal{L}=-\frac{1}{G}\sum_i\frac{1}{T_i}\sum_t\min\!\left(\rho_{i,t}A_i,\nobreak\operatorname{clip}(\rho_{i,t},1-\varepsilon_{\mathrm{clip}},1+\varepsilon_{\mathrm{clip}})A_i\right)+\beta\frac{1}{G}\sum_i\frac{1}{T_i}\sum_t d^{\mathrm{K3}}_{i,t}.
\end{equation}
The policy-clipping parameter is $\varepsilon_{\mathrm{clip}}=0.2$, and the K3 KL penalty coefficient is $\beta=0.03$.
All sums are masked to generated question tokens; prompt, dialogue, and image-token positions do not contribute to the policy loss.

\paragraph{Closed-source Luna baseline.} GPT-5.6 Luna is used only as an API questioner. It receives the ordered Top-4 candidate images and dialogue context, while the retriever, answerer, invalid-question gate, and five-round evaluation protocol remain identical to the main comparison. We use \texttt{reasoning\_effort=none}, temperature 1.0, top-p 0.5, \texttt{max\_completion\_tokens=100}, and up to 16 concurrent requests. The complete prompt and input format are given in Table~\ref{tab:prompt-luna-questioner}.

\paragraph{Invalid-question rules.} The validity gate is applied to a normalized, lower-cased question before the answerer and retriever in both training and test-time rollouts. It enforces a common answerable witness protocol: candidate indices, ranking instructions, self-filled answers, and memory-dump requests cannot enter the retrieval text. The following rules are applied in order:

\noindent\textbf{(1) Surface form.} Empty output, fewer than five words, or no question mark.

\noindent\textbf{(2) Selection and self-filling.} Candidate/image or ranking references, self-filled ``conversation'' or ``answer'' content, and unconditional requests for a complete memory or description.

\noindent\textbf{(3) Broad appearance and repetition.} Whole-person prompts or broad requests for additional details without a concrete local target such as clothing, bags, shoes, hair, accessories, body regions, colors, environment, setting, or background are rejected. A token-set Jaccard similarity of at least \(0.70\) with a previous dialogue question defines a historical repeat; historical repeats are invalid under the gate and receive the fixed invalid reward \(-0.2\).

\paragraph{State sampling details.} A state is identified by its query and interaction round. We remove states whose target is already Rank-1 before the action, sample across all valid rounds, and prevent duplicate states within a training pool. The source-dataset allocation is fixed to the proportions of Interactive-PEDES before random sampling. The 3K primary pool and 5K scaling pool use the same balanced construction procedure; only the number of sampled states changes.

\paragraph{Evaluation.} We use the 7,373-query Interactive-PEDES test split and a fixed gallery. We report Rank-1, Rank-5, Rank-10, mAP, and BRI after five rounds. All reported runs use fixed checkpoint paths and decoding parameters.

\paragraph{Question-type classification.} Figure~5 uses a deterministic lexical classifier based on question text. Each generated question is lower-cased, stripped, and flattened across line breaks. The rules are applied in this order: (i) a multiple-choice marker such as ``A)''--``E)'' yields the multiple-choice class; (ii) a question beginning with ``is,'' ``are,'' ``was,'' ``were,'' ``does,'' ``do,'' ``did,'' ``has,'' ``have,'' or ``had'' yields the yes/no class; (iii) broad prompts containing phrases such as ``any additional details,'' ``anything else,'' or ``appearance or surroundings'' yield the broad-open class; (iv) a question beginning with ``what,'' ``which,'' ``where,'' or ``how,'' or a request to describe/provide/tell an attribute such as clothing, shoes, bags, hair, hats, accessories, posture, or environment, yields the local-open class; and (v) all remaining questions yield the other class. Figure~5 aggregates the local-open bucket as local WH/open and the yes/no bucket as local yes/no. The classifier uses only the question text, not the answer, rank change, or reward, so the reported type proportions and conditional gains do not use downstream outcomes as labels.

\begin{figure}[H]
\centering
\begin{minipage}[t]{0.48\textwidth}
  \centering
  \includegraphics[width=0.95\linewidth]{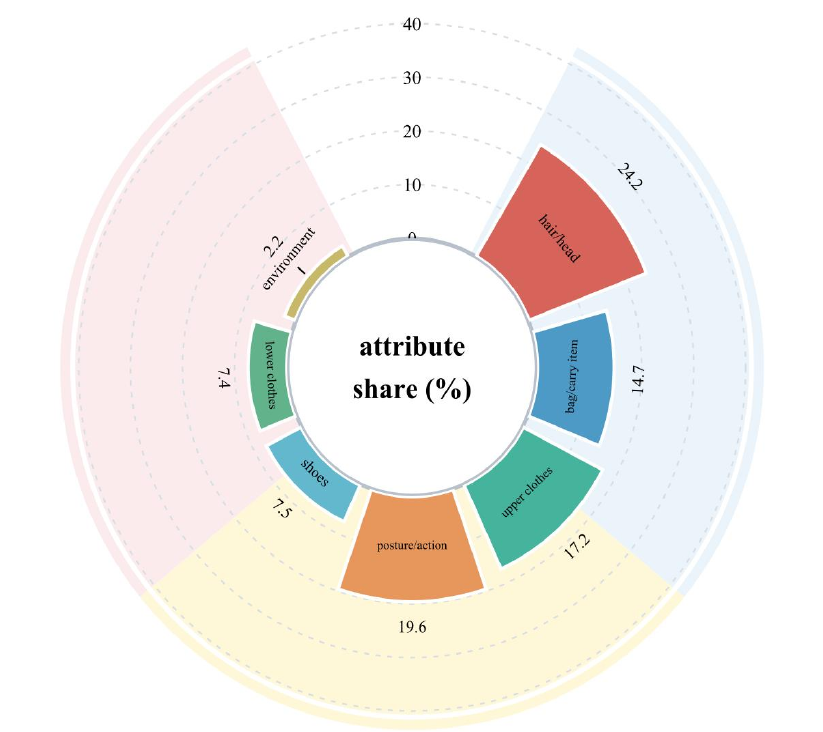}\\[-1pt]
  (a) LLaVA-ReID
\end{minipage}\hfill
\begin{minipage}[t]{0.48\textwidth}
  \centering
  \includegraphics[width=0.95\linewidth]{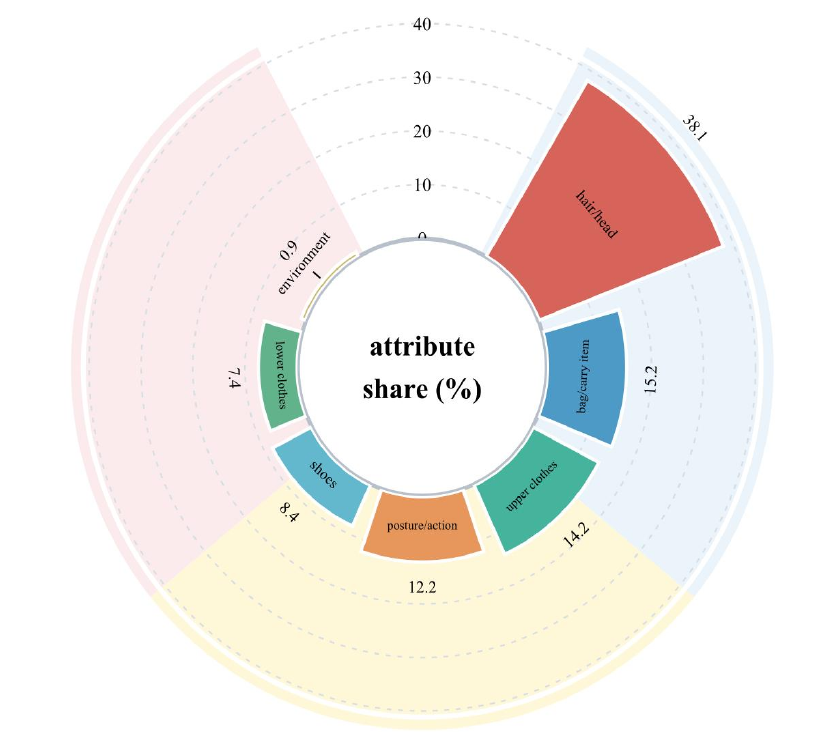}\\[-1pt]
  (b) RAVEL
\end{minipage}
\caption{Attribute shares before and after retrieval-aware training.}
\label{fig:supp-attribute-share}
\end{figure}

\noindent\makebox[\textwidth][c]{%
\begin{minipage}{0.75\textwidth}
\centering
\captionsetup{width=\linewidth}
\captionof{table}{Prompt used by GPT-5.6 Luna.}
\label{tab:prompt-luna-questioner}
\includegraphics[width=\linewidth]{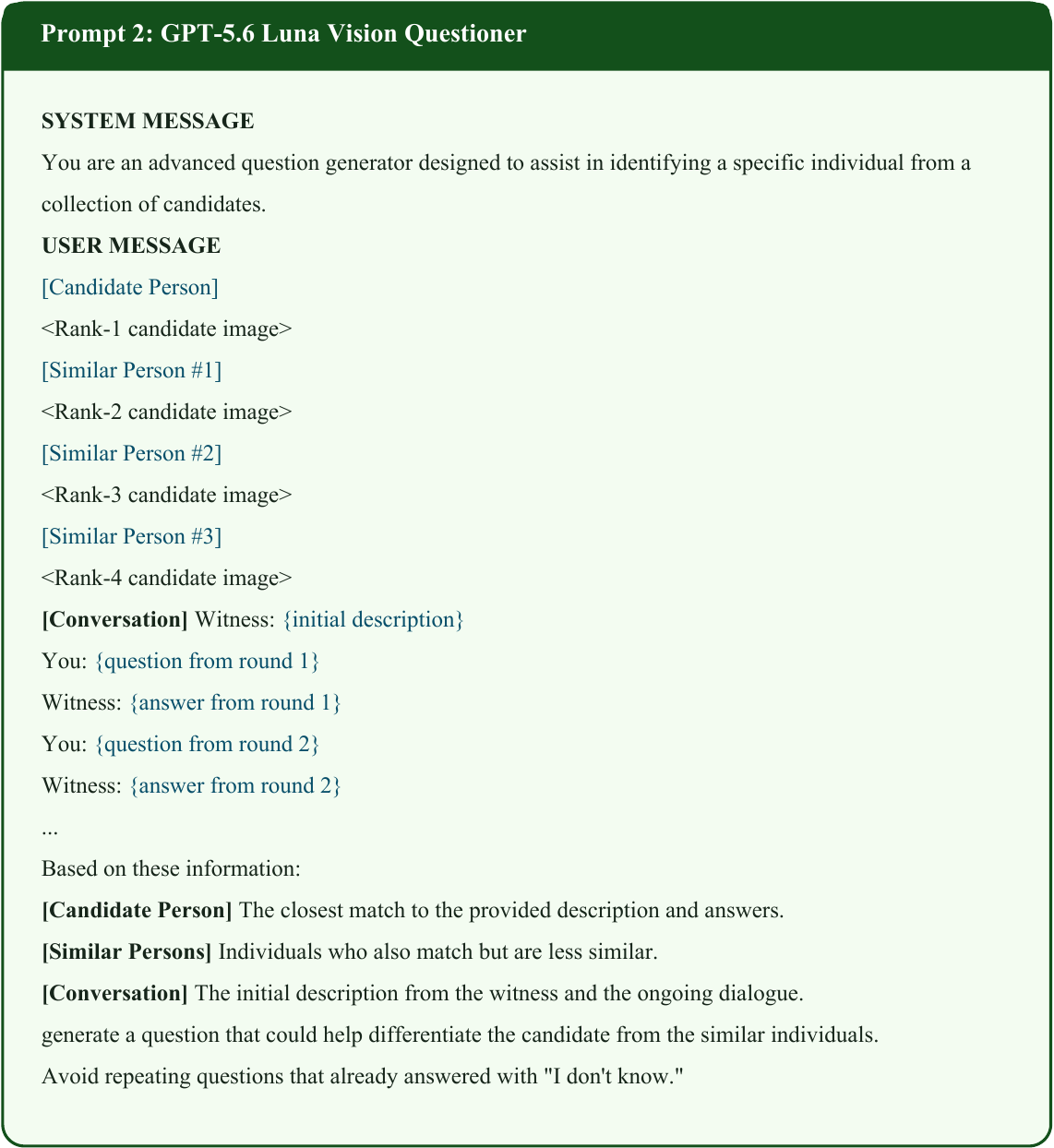}
\end{minipage}
}%

\clearpage
\section*{C. Qualitative Cases}
\noindent
To illustrate how retrieval-aware question learning changes the interaction trajectory, we present two representative five-round cases. Both methods receive the same initial description and candidate state; each panel then records the Top-4 candidates, the generated question, the witness answer, and the rank transition. The first case starts at Rank 5 and ends at Rank 3 for LLaVA-ReID versus Rank 1 for RAVEL, while the second case starts at Rank 52 and ends at Rank 85 versus Rank 1, respectively.
\begin{figure}[H]
\centering
\includegraphics[page=1,trim=150 495 150 10,clip,width=0.67\textwidth]{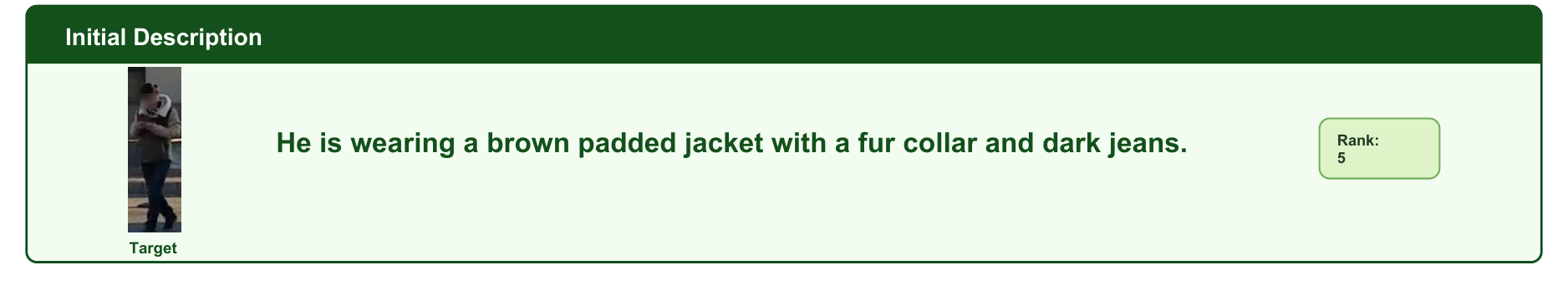}
\par
\includegraphics[page=1,trim=0 20 605 175,clip,width=0.74\textwidth]{figures_main_analysis/qualitative_cases.pdf}
\par\vspace{-4pt}
\includegraphics[page=1,trim=600 20 15 175,clip,width=0.74\textwidth]{figures_main_analysis/qualitative_cases.pdf}
\par
\caption{Qualitative comparison on a case where the target starts at Rank 5.}
\label{fig:qual-case-1}
\end{figure}
\clearpage
\begin{figure}[H]
\centering
\includegraphics[page=2,trim=150 495 150 10,clip,width=0.67\textwidth]{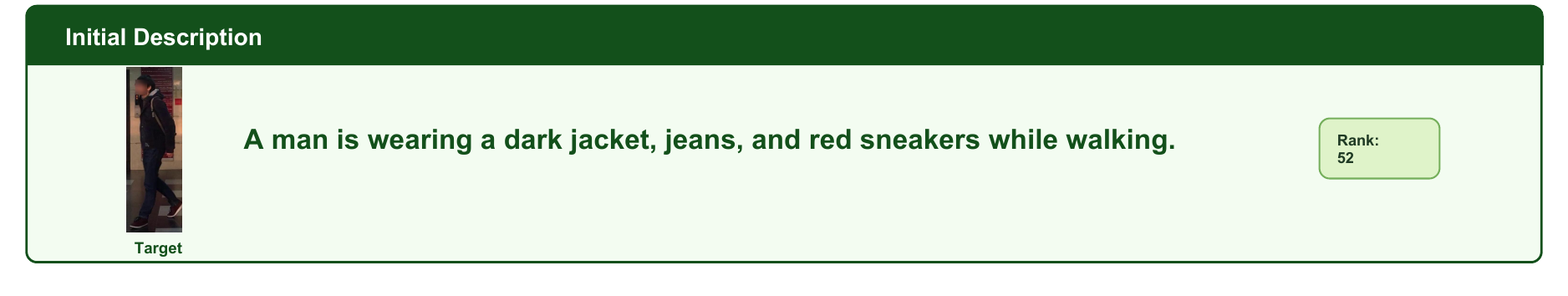}
\par
\includegraphics[page=2,trim=0 20 605 175,clip,width=0.84\textwidth]{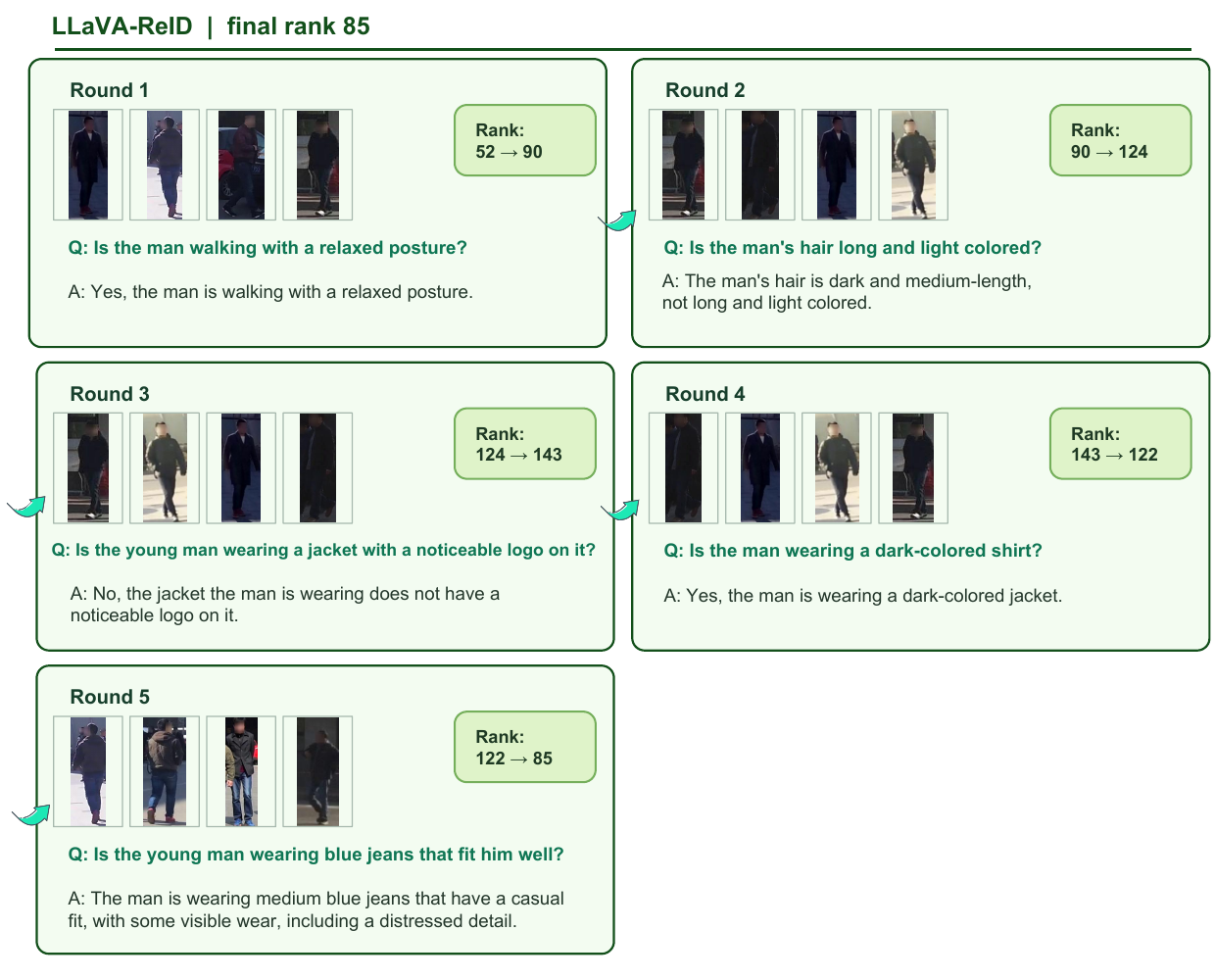}
\par\vspace{-4pt}
\includegraphics[page=2,trim=600 20 15 175,clip,width=0.84\textwidth]{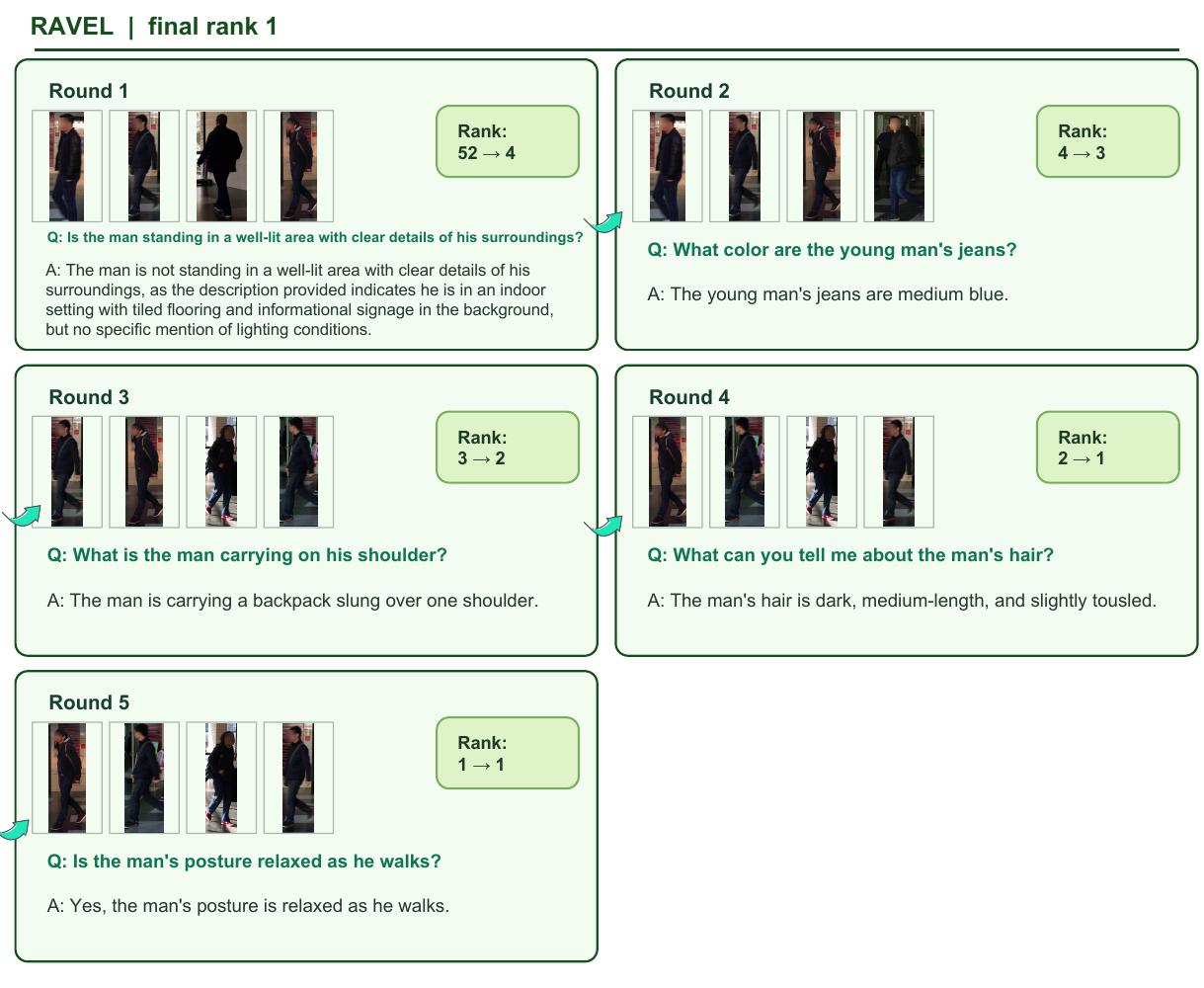}
\par
\caption{Qualitative comparison on a case where the target starts at Rank 52.}
\label{fig:qual-case-2}
\end{figure}

\end{document}